\documentclass[10pt, a4paper]{article}

\usepackage[final]{lrec-coling2024} 
\usepackage{inconsolata}
\usepackage{algorithm}
\usepackage{algorithmic}
\usepackage{amssymb}
\usepackage{dsfont}
\usepackage{amsmath}
\usepackage{booktabs}
\usepackage{multirow}
\usepackage[normalem]{ulem}
\useunder{\uline}{\ul}{}
\usepackage{amssymb}
\usepackage{bbm}
\usepackage{graphicx}
\usepackage{colortbl}
\usepackage{amsthm}
\usepackage{mathrsfs}

\title{UniPSDA: Unsupervised Pseudo Semantic Data Augmentation for Zero-Shot Cross-Lingual Natural Language Understanding}

\name{Dongyang Li$^{1,2}$,  {\bf \large Taolin Zhang$^{2}$,}  {\bf \large Jiali Deng$^{1}$,}  {\bf \large Longtao Huang$^2$,} \\ {\bf \large Chengyu Wang$^{2}$,}  {\bf \large Xiaofeng He$^{1}$,\thanks{Work done when Dongyang Li was doing an internship at Alibaba Group. Dongyang Li and Taolin Zhang contributed equally to this work. Correspondence to Chengyu Wang and Xiaofeng He.}}  {\bf \large Hui Xue$^2$}}

\address{$^1$School of Computer Science and Technology, East China Normal University \\$^2$Alibaba Group \\
         \{dongyangli0612, jialideng1127\}@gmail.com, hexf@cs.ecnu.edu.cn\\
         \{zhangtaolin.ztl, kaiyang.hlt, chengyu.wcy, hui.xueh\}@alibaba-inc.com\\}

\abstract{
Cross-lingual representation learning transfers knowledge from resource-rich data to resource-scarce ones to improve the semantic understanding abilities of different languages.
However, previous works rely on shallow unsupervised data generated by token surface matching, regardless of the global context-aware semantics of the surrounding text tokens.
In this paper, we propose an \textbf{Un}superv\textbf{i}sed \textbf{P}seudo \textbf{S}emantic \textbf{D}ata \textbf{A}ugmentation (UniPSDA) mechanism for cross-lingual natural language understanding to enrich the training data without human interventions.
Specifically, to retrieve the tokens with similar meanings for the semantic data augmentation across different languages, we propose a sequential clustering process in 3 stages: within a single language, across multiple languages of a language family, and across languages from multiple language families.
Meanwhile, considering the multi-lingual knowledge infusion with context-aware semantics while alleviating computation burden, we directly replace the key constituents of the sentences with the above-learned multi-lingual family knowledge, viewed as pseudo-semantic.
The infusion process is further optimized via three de-biasing techniques without introducing any neural parameters.
Extensive experiments demonstrate that our model consistently improves the performance on general zero-shot cross-lingual natural language understanding tasks, including sequence classification, information extraction, and question answering.
\\ \newline  }

\begin{document}

\maketitleabstract

\section{Introduction}

Cross-lingual representation learning facilitates resource-rich information to boost the performance of under-resourced languages in various downstream natural language understanding (NLU) tasks, such as text classification \cite{DBLP:conf/naacl/Huang22,DBLP:journals/kais/RathnayakeSRR22,DBLP:conf/rep4nlp/LiSZLR21}, sentiment analysis \cite{DBLP:conf/icdm/SzolomickaK22,DBLP:conf/aclnut/Sazzed20}, information extraction \cite{DBLP:conf/acl/HuangHNCP22,DBLP:conf/aaai/AhmadPC21,DBLP:conf/www/WangFHZ19,DBLP:conf/sdm/FanWCHH19}, and question answering \cite{DBLP:conf/naacl/LimkonchotiwatP22,DBLP:conf/www/PerevalovBDN22}.
Although existing cross-lingual works \cite{DBLP:conf/acl/LiHZDLHJWDZ23,DBLP:conf/emnlp/ClouatrePZC22} share explicit language semantics across different languages, they generally rely on supervised parallel corpora and simple, shallow unsupervised mechanisms such as back translation \cite{DBLP:conf/acl/LamSR22,DBLP:conf/acl/NishikawaRT21} and random deletion \cite{DBLP:conf/ijcai/0013GM0WW22}.

\begin{figure*}[!ht]
\centering
\includegraphics[width=0.9\textwidth]{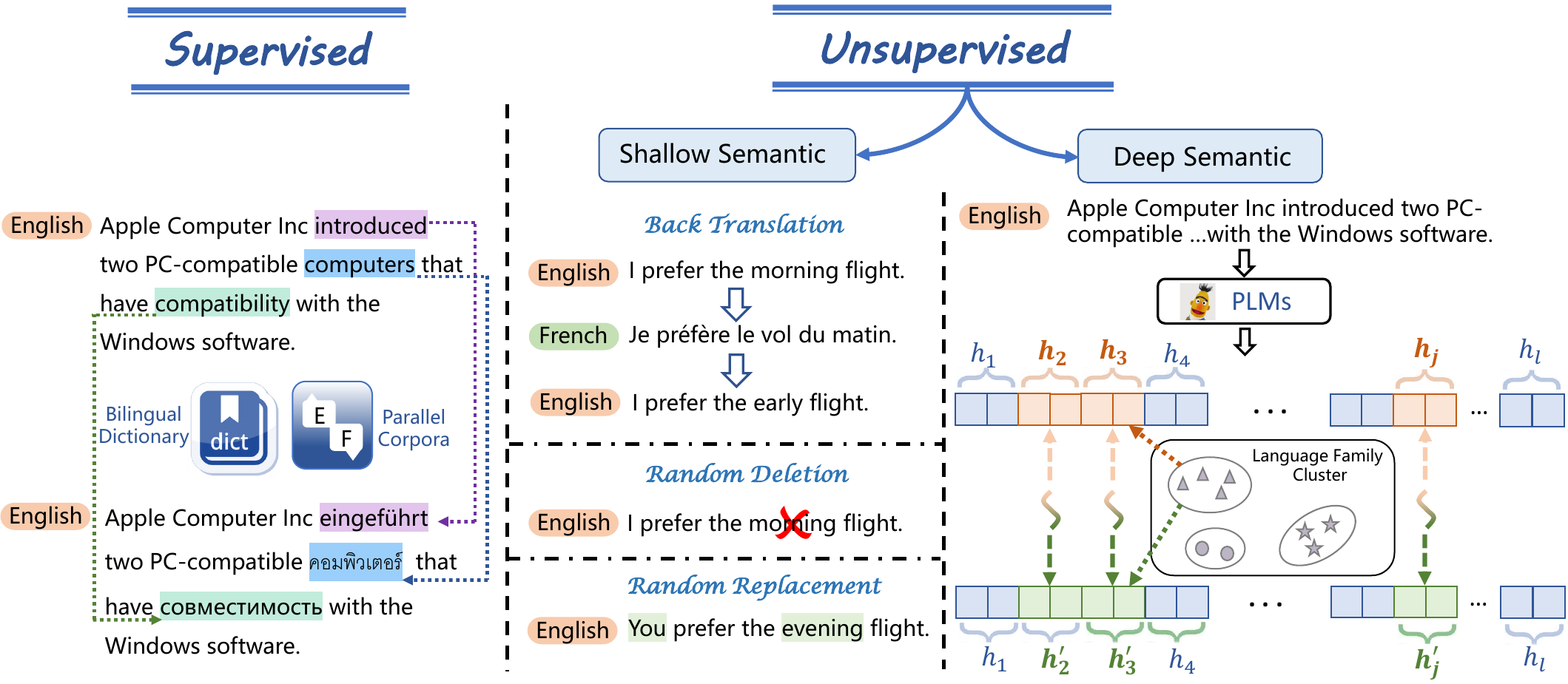}
\caption{Examples of previous data augmentation techniques, including supervised methods that rely on parallel data, and unsupervised methods which carry the risk of losing sentential semantic coherence.}
\label{introduction}
\end{figure*}

The previous data augmentation (DA) approaches in cross-lingual representation learning can be roughly divided into two categories: supervised parallel data augmenters and unsupervised shallow data augmenters.
\begin{enumerate}
    \item \emph{Supervised Parallel Data Augmenter:}
    These works \cite{DBLP:journals/kais/FernandoRSPR23,DBLP:conf/acl/LaiTN22,DBLP:conf/emnlp/RiabiSKSSS21} utilize annotated parallel corpora (e.g., bilingual dictionaries and translation tools) to augment the training data by aligning the same meanings across different languages for low-resource tasks. However, the collection process for these parallel corpora is time-consuming and relies on human annotation.
    \item \emph{Unsupervised Shallow Data Augmenter:}
    Unlike the supervised approaches mentioned above, these methods employ unsupervised easy data augmentation (EDA) techniques (e.g., back translation, random deletion, and random replacement) to generate additional training samples for model training \cite{DBLP:conf/acl/NishikawaRT21,DBLP:conf/acl/BariMJ20,DBLP:conf/acl/ChenKNGZOM20}.
    These methods focus solely on the surface semantics of the input samples to match cross-lingual data without considering the deeper linguistic connections.
\end{enumerate}
As shown in Figure \ref{introduction}, techniques like ``random deletion'' and ``random replacement'' may alter the sentence's intended meaning. Hence, we aim to expand the multilingual training samples based on a deep semantic understanding that the model can automatically derive, such as from the hidden layers of a pre-trained language model (PLM).



To overcome the issues mentioned above, we propose an \textbf{Un}superv\textbf{i}sed \textbf{P}seudo \textbf{S}emantic \textbf{D}ata \textbf{A}ugmentation (UniPSDA) mechanism, which mainly consists of two modules:
\begin{itemize}
    \item \textbf{Domino Unsupervised Cluster:} 
    To provide high-quality multilingual representations for performing the subsequent deep unsupervised data augmentation, we group languages into a hierarchical structure organized by language families\footnote{\url{https://www.ethnologue.com/browse/families}} to learn multilingual relations. 
    We perform the clustering process via the domino chain process\footnote{\url{https://en.wikipedia.org/wiki/Domino_effect}} to collect semantically similar words across different languages by comparing the embeddings themselves, a method we name Domino Unsupervised Cluster.
    Specifically, the domino cluster is a chain-rule process comprised of three different sequential stages: the single language stage, the language family stage, and the multi-language stage.
    \item \textbf{Pseudo Semantic Data Augmentation:} Considering that previous data augmentation methods focus on the surface of naive training samples, we employ the learned multilingual internal representations to address the semantic deficiencies of the training samples.
    Specifically, the domino clustering-enhanced ultimate multilingual representations directly replace the important positions' hidden states in training samples, as recognized by the \texttt{<subject,verb,object>} (SVO) structure.
    The potential incompatibility phenomena of inserting clustering multilingual representations may result in biased parameter learning.
    To further alleviate the misalignment between the replaced embeddings space and the context output space of PLMs, we introduce three de-biasing optimal transport affinity regularization techniques to make the learning process faster and more stable.
\end{itemize}



\section{Methodology}
\subsection{Model Notations}

\begin{figure*}[tb]
\centering
\includegraphics[width=0.95\textwidth]{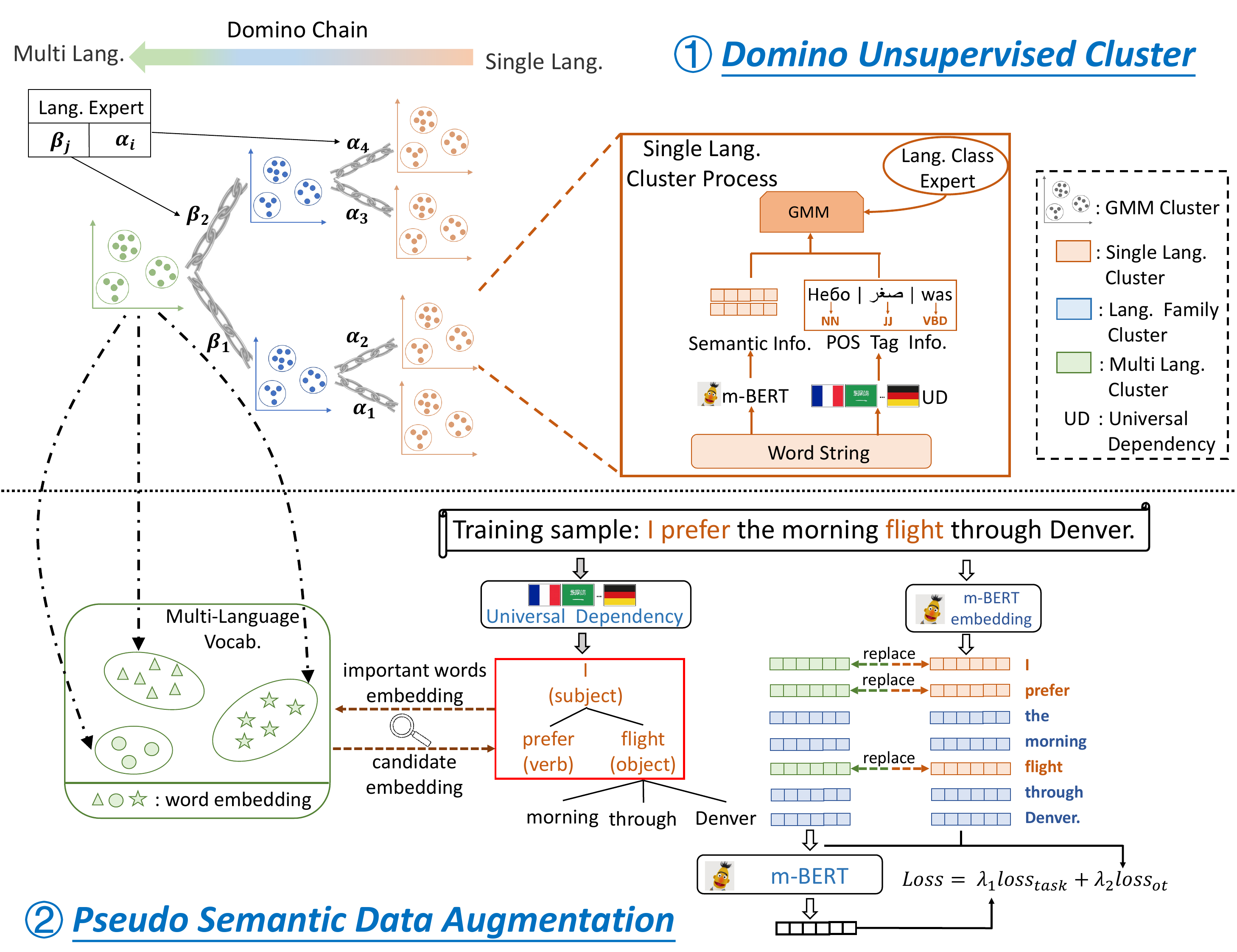} 
\caption{Model Overview of UniPSDA. (Best viewed in color)}
\label{overview_figure}
\end{figure*}
The architecture of UniPSDA is shown in Figure~\ref{overview_figure}.
The goal of cross-lingual natural language understanding is to utilize a source language dataset $\mathcal{D}_{\text{lang}}= (\mathcal{X}_{\text{lang}},\mathcal{Y}_{\text{lang}})$ to train a model $\mathcal{M}$.
Then we apply the trained model $\mathcal{M}$ to tasks in other target languages $\mathcal{D}_{\text{lang}'}= (\mathcal{X}_{\text{lang}'},\mathcal{Y}_{\text{lang}'})$, where $\mathcal{X}$ denotes the input samples and $\mathcal{Y}$ is the label set.
In our work, each sentence of the training data is denoted as $S_i =  ( w_{i1}, w_{i2}, \cdots, w_{ij}, \cdots, w_{il_i} )$, where $w_{ij}$ denotes the $j$-th word in sentence $S_i$ and $l_i$ is the maximum word count of the sentence. The hidden state of word $w_{ij}$ is $h_{w_{ij}}\in \mathbb{R}^{|u|\times d}$, where $|u|$ is the maximum number of tokens contained in the word and $d$ is the dimension of the hidden state. The hidden state of sentence $S_i$ is $h_{s_i}\in \mathbb{R}^{|L_s|\times d}$, where $|L_s|$ is the sentence's maximum token length.
The specific notations for the three clustering stages in the \texttt{Domino Unsupervised Cluster} are as follows:
\begin{itemize}
    \item In the single language stage, the words in the $m$-th single language $G_{\text{Sin}_m}$ are clustered into $|G_{\text{Sin}_m}|$ clusters. The $t$-th cluster is denoted as $Clu_{mt}^{\text{sin}}$.
    \item In the language family stage, the words in the $n$-th language family $G_{\text{Fam}_n}$ are clustered into $|G_{\text{Fam}_n}|$ clusters. The $g$-th cluster is denoted as $Clu_{ng}^{\text{fam}}$.
    \item In the multi-language stage, all language families are collected into $G_{\text{Mul}}$. All the words in $G_{\text{Mul}}$ are clustered into $|G_{\text{Mul}}|$ clusters. The $q$-th cluster is denoted as $Clu_q^{\text{mul}}$.
\end{itemize}

\subsection{Text Encoder}
In this paper, m-BERT \cite{DBLP:conf/naacl/DevlinCLT19} is utilized as our encoder\footnote{Other multilingual pre-trained language models can also be considered as the backbone.} to obtain the hidden states, which are averaged from the embeddings of the first and last layers. 
The final hidden state of the $j$-th word in sentence $S_i$ is formulated as:
\begin{equation}
     h_{w_{ij}} = \frac{1}{2} \left ( \mathcal{F}_{\text{first}} \left(w_{ij}\right) + \mathcal{F}_{\text{last}}\left(w_{ij}\right)   \right )
\end{equation}
where $\mathcal{F}_{\text{first}}$ and $\mathcal{F}_{\text{last}}$ denote the representations from the first and last layers, respectively. We average these representations to obtain the sentence's hidden state $h_{s_i}$.


\subsection{Domino Unsupervised Cluster}
To enable the model to learn relevant word information corresponding to different languages, we perform three hierarchical, chain-rule-based clustering steps, sequentially applied to representations of varying language granularities.




\subsubsection{Single Language Cluster}
In the single language cluster stage, we aim to group similar words within a specific language. 
To refine the clustering process, we clarify that ``similar words'' refers not only to semantic similarity but also to the concordance of part-of-speech (POS) tags. 
For instance, verbs with the meaning of ``hope'' are grouped together, distinct from nouns with similar meanings, thereby incorporating lexical POS knowledge into the clustering.
Initially, we employ the Universal Dependencies\footnote{\url{https://universaldependencies.org/}}-based PyTorch tool Stanza\footnote{Stanza is an off-the-shelf cross-lingual linguistic analysis package. URL:~\url{https://stanfordnlp.github.io/stanza/}}, to obtain the POS tag for each word.
The training data contains 17 types of POS tags, and we represent each word with a 17-dimensional one-hot vector $v_{_{\text{POS}}}$ to signify the initial tag representations.
This one-hot vector is then mapped to a context-aware space by a linear function $(W_{_{\text{POS}}}v_{_{\text{POS}}} + b_{_{\text{POS}}})$.
The final embeddings $h_{w_{ij}}^{\text{final}}$ are obtained by concatenating the original word representations with the projected POS tags:
\begin{equation}
     h_{w_{ij}}^{\text{final}} = [h_{w_{ij}} \; || \; (W_{_{_{\text{POS}}}}v_{_{_{\text{POS}}}} + b_{_{_{\text{POS}}}})]
\end{equation}
where ``$||$'' denotes the concatenation operation.
Words with similar final embeddings are clustered together using an expectation-maximization algorithm based on Gaussian Mixture Models (GMM).
For the $m$-th language, we obtain $|G_{\text{Sin}_m}|$ clusters at the end of the clustering process:
\begin{equation}
\small
\begin{split}
    Clu_{m1}^{\text{sin}},\; Clu_{m2}^{\text{sin}},\; \ldots, \;Clu_{m|G_{\text{Sin}_m}|}^{\text{sin}} = \text{GMM}(\\ h_{w_{1}}, h_{w_{2}}, \ldots, h_{w_{|G_{\text{Sin}_m}|}})
\end{split}
\end{equation}
where $|G_{\text{Sin}_m}|$ is the total number of words in the $m$-th language.

\subsubsection{Language Family Cluster}
\label{section_language_family_cluster}
In the Ethnologue\footnote{\url{https://www.ethnologue.com/}} linguistic categorical tree, each language is considered a leaf node.
Language families serve as ancestor nodes within this tree structure, and all descendant nodes of a particular ancestor node are grouped into the same language family. 
We aggregate the results of single language clusters within a specific language family and calculate the expert weight $\alpha_{mt}$ for each cluster using a Gate mechanism \cite{DBLP:conf/emnlp/LiLQ18}.
This weight is determined by the proportion of each cluster's word count in relation to the total sample size. In essence, we incorporate the size information of each cluster into the clustering process. The single language stage involves a total of $N_{\text{sin}} = |Clu_{m1}^{\text{sin}}| + |Clu_{m2}^{\text{sin}}| + \cdots + |Clu_{m|G_{\text{Sin}_m}|}^{\text{sin}}|$ elements, where $|Clu_{m1}^{\text{sin}}|$ denotes the number of elements in cluster $Clu_{m1}^{\text{sin}}$.
Thus, the expert weight for the $t$-th cluster $\alpha_{mt}$ of language $G_{\text{Sin}_m}$ is defined as $\alpha_{mt} = \frac{|Clu_{mt}^{\text{sin}}|}{N_{\text{sin}}}$.
We represent all expert weights across $r$ languages of the $n$-th language family $G_{\text{Fam}_n}$ as the matrix $A_{\text{sin}}$.
The cluster-center embeddings of single language clusters, denoted as $Cen_{mt}^{\text{sin}}$, are used in the language family cluster stage. 
The expert-weighted cluster-center embeddings $h_{Cen_{mt}^{\text{sin}}}\in \mathbb{R}^{|u|\times d}$ are then input into the GMM clustering process.
The matrix $H_{\text{sin}}$ comprises all cluster-center embeddings across $r$ languages of the $n$-th language family $G_{\text{Fam}_n}$.
GMM clustering groups semantically similar words from $r$ languages into a specific cluster $Clu_{ng}^{\text{fam}}$.
\begin{equation}
\begin{aligned}
Clu_{n1}^{\text{fam}}, Clu_{n2}^{\text{fam}}, \ldots, Clu_{n|G_{\text{Fam}_n}|}^{\text{fam}} = \text{GMM}(\\ \text{ele}(A_{\text{sin}} \odot H_{\text{sin}}))
\end{aligned}
\end{equation}
where the $A_{\text{sin}}$ and $H_{\text{sin}}$ matrices are defined as follows:
\begin{equation}
\begin{aligned}
A_{\text{sin}} &= \begin{bmatrix}
\alpha_{11} & \alpha_{12} & \cdots & \alpha_{1|G_{\text{Sin}_1}|} \\
\alpha_{21} & \alpha_{22} & \cdots & \alpha_{2|G_{\text{Sin}_2}|} \\
\vdots & \vdots & \ddots & \vdots \\
\alpha_{r1} & \alpha_{r2} & \cdots & \alpha_{r|G_{\text{Sin}_r}|}
\end{bmatrix}, \\
H_{\text{sin}} &= \begin{bmatrix}
h_{Cen_{11}^{\text{sin}}} & h_{Cen_{12}^{\text{sin}}} & \cdots & h_{Cen_{1|G_{\text{Sin}_1}|}^{\text{sin}}} \\
h_{Cen_{21}^{\text{sin}}} & h_{Cen_{22}^{\text{sin}}} & \cdots & h_{Cen_{2|G_{\text{Sin}_2}|}^{\text{sin}}} \\
\vdots & \vdots & \ddots & \vdots \\
h_{Cen_{r1}^{\text{sin}}} & h_{Cen_{r2}^{\text{sin}}} & \cdots & h_{Cen_{r|G_{\text{Sin}_r}|}^{\text{sin}}}
\end{bmatrix}
\end{aligned}
\end{equation}
where $\text{ele}()$ denotes the operation of enumerating every element of the matrix, and $\odot$ represents the element-wise product.


\subsubsection{Multi Languages Cluster}
Finally, we perform clustering on all language family cluster-center embeddings obtained from the language family cluster stage. For example, the first cluster-center of cluster $Clu_{n1}^{\text{fam}}$ in the $n$-th language family is denoted as $Cen_{n1}^{\text{fam}}$.
Each cluster-center embedding is associated with a multi-language expert weight $\beta_{ng}$, computed using the same Gate mechanism as in the language family cluster stage.
We represent all expert-weight elements across $z$ language families of the multi-language pool $G_{\text{Mul}}$ as the matrix $B_{\text{fam}}$.
These expert-weighted cluster-center embeddings $h_{Cen_{ng}^{\text{fam}}} \in \mathbb{R}^{|u| \times d}$ are then used in the GMM clustering process.
The matrix $H_{\text{fam}}$ contains all cluster-center embeddings across the $z$ language families of $G_{\text{Mul}}$.
GMM clustering groups semantically similar words from the $z$ language families into specific clusters, denoted as $Clu_{q}^{\text{mul}}$.
\begin{equation}
\begin{aligned}
Clu_{1}^{\text{mul}}, Clu_{2}^{\text{mul}}, \ldots, Clu_{|G_{\text{Mul}}|}^{\text{mul}} = \text{GMM}(\\ \text{ele}(B_{\text{fam}} \odot H_{\text{fam}}))
\end{aligned}
\end{equation}
where the matrices $B_{\text{fam}}$ and $H_{\text{fam}}$ are defined as follows:
\begin{equation}
\begin{aligned}
B_{\text{fam}} &= \begin{bmatrix}
\beta_{11} & \beta_{12} & \cdots & \beta_{1|G_{\text{Sin}_1}|} \\
\beta_{21} & \beta_{22} & \cdots & \beta_{2|G_{\text{Sin}_2}|} \\
\vdots & \vdots & \ddots & \vdots \\
\beta_{z1} & \beta_{z2} & \cdots & \beta_{z|G_{\text{Sin}_z}|}
\end{bmatrix}, \\
H_{\text{fam}} &= \begin{bmatrix}
h_{Cen_{11}^{\text{fam}}} & h_{Cen_{12}^{\text{fam}}} & \cdots & h_{Cen_{1|G_{\text{Fam}_1}|}^{\text{fam}}} \\
h_{Cen_{21}^{\text{fam}}} & h_{Cen_{22}^{\text{fam}}} & \cdots & h_{Cen_{2|G_{\text{Fam}_2}|}^{\text{fam}}} \\
\vdots & \vdots & \ddots & \vdots \\
h_{Cen_{z1}^{\text{fam}}} & h_{Cen_{z2}^{\text{fam}}} & \cdots & h_{Cen_{z|G_{\text{Fam}_z}|}^{\text{fam}}}
\end{bmatrix}
\end{aligned}
\end{equation}
where $\text{ele}()$ denotes the operation of enumerating each element of the matrix, and $\odot$ represents the element-wise product.



\subsection{Pseudo Semantic Data Augmentation}
To enrich the training data with diverse linguistic information, we augment the model with global multilingual semantics obtained from the last domino unsupervised cluster module.

\subsubsection{Pseudo Semantic Replacement}
We propose two approaches for handling sentence semantics. 
First, we pass sentences through m-BERT to generate embeddings for each sentence. 
Second, we use Universal Dependencies to extract syntactic parsing trees for the sentences.
To guide the model toward learning more accurate representations of crucial sentence words, we focus on key elements identified through syntactic parsing. 
Given that the subject, verb, and object (SVO) components are essential for comprehension in many tasks \cite{DBLP:conf/cvpr/DaiZL17,DBLP:conf/iccv/ZhangKYC17}, we treat the SVO as the crucial words of each sentence. 
Subsequently, we mark the position of each SVO component within the sentence:
\begin{equation}
\begin{split}
    h_{s_i} = [e_{t_{i1}}, e_{t_{i2}}, \ldots, e_{w_{iS}}, \ldots, \\ e_{w_{iV}}, \ldots, e_{w_{iO}}, \ldots, e_{t_{i|L_s|}} ]
\end{split}
\end{equation}
where $e_{w_{iS}}$, $e_{w_{iV}}$, and $e_{w_{iO}}$ denote the embeddings of the sentence's subject, verb, and object components, respectively, and $e_{t_{i1}}$ represents the embedding of the first token in sentence $S_i$. 
We replace the original sentence's SVO word embeddings with randomly selected candidate embeddings from the same cluster. These candidate word embeddings share similar semantics with the SVO words but come from different languages.
Through the replacement of crucial words with cross-lingual knowledge, we can guide the model to learn more about the critical linguistic elements of sentences and achieve better semantic representations. The updated sentence representation is expressed as:
\begin{equation}
\begin{split}
 h_{s_i}^{'} = [e_{t_{i1}}, e_{t_{i2}}, \ldots, e_{can_{iS}'}, \ldots,\\ e_{can_{iV}'}, \ldots, e_{can_{iO}'}, \ldots, e_{t_{i|L_s|}} ]   
\end{split}
\end{equation}
where $e_{can_{iS}'}$, $e_{can_{iV}'}$, and $e_{can_{iO}'}$ denote the candidate embeddings for the subject, verb, and object components, respectively.
After the embedding replacement, cross-lingual pseudo semantic information is introduced to the training data. We then feed these enhanced representations into transformer models with base-level parameter sizes to refine the embeddings.

\subsubsection{De-biasing Optimal Transport Affinity Regularization}
Spatial misalignment exists between the original sentence and the enhanced sentence, as noted by \cite{DBLP:conf/cvpr/HuangYZWQY22}.
To diminish the discrepancy between the replaced sentence embedding $h_{s_{i}}^{'}$ and the original sentence embedding $h_{s_{i}}$, we introduce an integrated regularization term based on the optimal transport mechanism, named Optimal Transport Affinity Regularization.

\textbf{(1) Wasserstein Distance Abbreviation:}
To align the space of original sentence representations with that of cross-lingual knowledge-enhanced sentence representations \cite{DBLP:conf/icassp/WangH22,DBLP:conf/emnlp/AlqahtaniLZRM21}, we employ optimal transport ($\mathcal{OT}$) to facilitate the adjustment process.
We calculate a transport plan $P$ that maps the original sentence to the augmented sentence with optimal cost $C \in \mathbb{R}^{|L_{s}|\times|L_{s}|}$, using the Euclidean distance \cite{danielsson1980euclidean} between the two sentence representations as a measure of cost:
\begin{equation}
\small
 C(h_{s_{i}},h_{s_{i}}^{'}) = \left( \sum_{j=1}^{d}{\left| h_{s_{i}j}-h_{s_{i}j}^{'} \right|^{2}} \right)^{\frac{1}{2}}
\end{equation}
We aim to find the optimal transport plan $P \in \mathbb{R}^{|L_{s}|\times|L_{s}|}$ that minimizes the cost $C$. This problem is formulated as minimizing the $p$-Wasserstein distance $d_{p-Wass}$. Due to the high computational complexity of calculating $P$, we approximate it using the Sinkhorn algorithm \cite{DBLP:conf/nips/AltschulerWR17}:
\begin{equation}
\small
\mathbf{K} = \exp\left( -\frac{C(h_{s_{i}},h_{s_{i}}^{'})}{\varepsilon} \right)  
\end{equation}
\begin{equation}
P(h_{s_{i}},h_{s_{i}}^{'}) = diag(u)\mathbf{K}diag(v)
\end{equation}
We compute $u$ and $v$ iteratively, starting with $v^{(0)}=\mathbf{1}_{|L_{s}|}$, using the following update formulas:
\begin{equation}
u^{(l+1)}=\frac{a(h_{s_{i}})}{\mathbf{K}v^{(l)}},\quad v^{(l+1)}=\frac{b(h_{s_{i}}^{'})}{\mathbf{K}^{T}u^{(l+1)}}
\end{equation}
where $a$ and $b$ are distribution mapping functions.
The $\mathcal{OT}$ loss can be defined as:
\begin{equation}
 loss_{\mathcal{OT}} = \langle P(h_{s_{i}},h_{s_{i}}^{'}), C(h_{s_{i}},h_{s_{i}}^{'}) \rangle
\end{equation}
To mitigate the $\mathcal{OT}$ learning biases between the two sentence representations, we introduce two auxiliary de-biasing terms to calibrate the loss.

\textbf{(2) De-biasing Eigenvectors Shrinkage:}
We utilize a linear orthogonal mapping parameter $\mathbf{W} \in \mathbb{R}^{|L_{s}|\times |L_{s}|}$ to approximate the replaced embeddings to the original ones, $h_{s_{i}} \approx \mathbf{W}h_{s_{i}}^{'}$. Singular value decomposition (SVD) is directly applied to compute $\mathbf{W}$ \cite{DBLP:conf/naacl/XingWLL15}:
\begin{equation}
\small
\mathbf{U \Sigma V^{T}} = \text{SVD}(h_{s_{i}}^{'T} h_{s_{i}})
\end{equation}
\begin{equation}
\small
\mathbf{W = VU^{T}}
\end{equation}
We initialize the linear mapping function with weight $\mathbf{W}$ to simplify the learning process. However, eigenvectors with small singular values can lead to poor transformations if not suppressed \cite{DBLP:conf/nips/ChenWFLW19}. Thus, we penalize the smallest $k$ singular values of $\Sigma$, which is ordered by magnitude. The eigenvectors shrinkage loss is defined as:
\begin{equation}
loss_{eig} = - \eta \sum_{r = 1}^{k}\sigma _{r}^{2}
\end{equation}
where $\eta$ is a hyper-parameter to control the degree of penalty, and $\sigma _{r}$ is the $r$-th smallest singular value.

\textbf{(3) De-biasing Distance Shrinkage:}
To guide the framework's learning direction towards minimizing the discrepancy, we add a term based on the distance between the two embeddings to the loss function. The distance shrinkage loss is defined as:
\begin{equation}
loss_{dis} = 1 - \text{sim}(h_{s_{i}}, h_{s_{i}}^{'})
\end{equation}
where $\text{sim}()$ represents the similarity measure.

Finally, the auxiliary $\mathcal{OT}$ affinity regularization is given by:
\begin{equation}
loss_{Reg} = \rho _{1}loss_{\mathcal{OT}} + \rho _{2}loss_{eig} + \rho _{3}loss_{dis}
\end{equation}
where $\rho_{i}$ denotes the controlled weight of each regularization component, with the constraint that the sum of $\rho_{i}$ equals 1.

\subsection{Training Objective}
Our training objective combines the task-specific loss with the $\mathcal{OT}$ affinity regularization. The overall objective function is formulated as:
\begin{equation}
loss_{total} = \lambda _{1}loss_{task} + \lambda _{2}loss_{Reg}
\end{equation}
where $\lambda _{i}$ controls the relative contribution of each component, and the sum of $\lambda _{i}$ is constrained to be 1.

\section{Experiments}


\subsection{Tasks and Datasets}
\label{appendix_sec:Tasks_Datasets}
\textbf{Sequence Classification} tasks include text classification and sentiment analysis.
We selected the following datasets for these tasks: MLDoc \cite{DBLP:conf/lrec/SchwenkL18} for text classification, and the MultiBooked Catalan and Basque \cite{DBLP:conf/lrec/BarnesBL18}\footnote{We refer to these datasets collectively by the term ``OpeNER''.} for sentiment analysis.
The evaluation metrics for these tasks are accuracy and macro F1.

For \textbf{Information Extraction}, we focus on Relation Extraction as a representative task. Here, the goal is to predict the correct relation type present in the data. We use the ACE2005 dataset \cite{walker2006ace}, which spans three languages: English, Chinese, and Arabic. The performance is measured using micro F1.

\textbf{Question Answering} involves retrieving answers for specific questions from a given passage. We conduct experiments on the cross-lingual question answering dataset BiPaR \cite{DBLP:conf/emnlp/JingXZ19}, which is commonly used for evaluating such systems. The evaluation metrics for this task are Exact Match (EM) and micro F1.


\begin{table*}[tb]
\centering
\begin{tabular}{c|cccccccc|c}
\toprule
\textbf{Model}                                               & \textbf{en}   & \textbf{de}   & \textbf{zh}   & \textbf{es}   & \textbf{fr}   & \textbf{it}   & \textbf{ja}   & \textbf{ru}   & \textbf{Average} \\ \midrule
MLDoc                                                                         & 87.2                           & 71.7                           & 73.5                           & 65.3                           & 70.2                           & 65.1                           & 69.8                           & 56.9                           & 69.9$_{(\pm 0.7)}$                              \\
LASER                                                                         & 86.5                           & 86.0                           & 70.4                           & 71.3                           & 73.9                           & 65.6                           & 58.5                           & 63.4                           & 72.0$_{(\pm 0.5)}$                              \\
m-BERT                                                                        & 92.1                           & 74.3                           & 72.5                           & 67.0                           & 70.5                           & 61.8                           & 69.7                           & 61.5                           & 71.2$_{(\pm 0.3)}$                              \\
XLM-R                                                                       &  90.7 & 78.5 & 70.3 & 66.4 & 67.8 & 63.9 & 64 & 64.0 & 70.7$_{(\pm 0.6)}$                              \\
ZSIW                                                                          &  91.3    &  82.8    &  79.6    &  71.7    &  78.1    &  67.0    &  68.5    &  64.3    & 75.4$_{(\pm 0.5)}$                              \\
DAP                                                                           &  94.1    &  86.7    &  81.7    &  76.2    & 84.3                           &  67.6    &  73.9    &  66.7    &  78.9$_{(\pm 0.2)}$      \\
SOGO$_{cos}$                                                                           &93.2 & 87.0 & 81.8 & 76.2 & 82.5 & 68.7 & 73.7 & 63.9 & 78.4$_{(\pm 0.1)}$  \\
X-STA                                                                           &  93.8 & 86.4 & 81.7 & 77.2 & 84.3 & 68.4 & 73.4 & 64.8 & 78.8$_{(\pm 0.2)}$  \\
CoSDA-ML                                                                      & 92.4                           & 79.1                           & 72.7                           & 69.9                           & 74.5                           & 64.3                           & 70.6                           & \textbf{66.9} &  73.8$_{(\pm 0.6)}$                              \\ \midrule
\rowcolor[HTML]{C0C0C0}\textbf{UniPSDA} & \textbf{94.5} & \textbf{87.1} & \textbf{82.3} & \textbf{77.4} & \textbf{84.4} & \textbf{69.4} & \textbf{74.0} & 65.5                           &  \textbf{79.3}$_{(\pm 0.2)}$    \\ \bottomrule
\end{tabular}
\caption{General results of text classification in terms of accuracy (\%) on the MLDoc dataset.}
\label{text_classification_general_result}
\end{table*}

\subsection{Experiment Settings}
\label{appendix_sec:Experiments_Settings}
Given computational resource constraints, we employ the base-level version of multilingual BERT (m-BERT) to obtain hidden states for words and sentences. The encoder consists of 12 Transformer layers with 12 self-attention heads, and the hidden state dimension is set to 768. During training, we experiment with learning rates in \(\{1\text{e}-5, 2\text{e}-5, 3\text{e}-5, 1\text{e}-6, 2\text{e}-6, 3\text{e}-6\}\). AdamW is chosen as the optimizer, with a learning rate of \(1\text{e}-3\) and weight decay of \(1\text{e}-5\). For the Wasserstein distance, we set \(p=1\), while the Sinkhorn algorithm's control parameter \(\varepsilon\) is \(0.1\). The last \(k=300\) singular values are used in the De-biasing Eigenvectors Shrinkage section, with \(\eta\) in the \(loss_{\text{eig}}\) formula being \(0.001\). The weight \(\rho\) of \(loss_{\text{Reg}}\) is set to \(\{0.4, 0.2, 0.4\}\), and the \(\lambda\) of the total loss is set to \(\{0.5, 0.5\}\) independently. Statistical results are based on 5 runs, and t-tests confirm that improvements are statistically significant, with \(p<0.05\) for all results.\footnote{The source code and data are available at \url{https://github.com/MatNLP/UniPSDA}}

\subsection{Baselines}
\label{appendix_sec:Baselines}
We compare our approach against a variety of baselines:

\textbf{MLDoc} \cite{DBLP:conf/lrec/SchwenkL18} introduces a cross-lingual text classification dataset, with baseline results from basic neural network models.

\textbf{BLSE} \cite{DBLP:conf/acl/BarnesWK18} presents a model for the sentiment analysis task, relying on supervised parallel bilingual data.

\textbf{LASER} \cite{DBLP:journals/tacl/ArtetxeS19} proposes a system utilizing a BiLSTM to learn sentence representations across 93 languages, assessed on natural language understanding (NLU) tasks.

\textbf{m-BERT} \cite{DBLP:conf/naacl/DevlinCLT19} offers a language model pre-trained on over 100 languages, generating representations for different languages.

\textbf{XLM-R} \cite{DBLP:conf/acl/ConneauKGCWGGOZ20} is a transformer-based masked language model known for its strong cross-lingual performance.

\textbf{mUSE} \cite{DBLP:conf/acl/YangCAGLCAYTSSK20} is pre-trained in 16 languages to project multilingual corpora into a single semantic space.

\textbf{CoSDA-ML} \cite{DBLP:conf/ijcai/QinN0C20} proposes a model using shallow string surface data augmentation to include various language strings in the training data.

\textbf{CCCAR} \cite{DBLP:conf/emnlp/NguyenNMN21} designs a model for information extraction tasks, leveraging datasets in three target languages.

\textbf{ZSIW} \cite{DBLP:conf/rep4nlp/LiSZLR21} introduces a zero-shot instance-weighting model for cross-lingual text classification.

\textbf{HERBERTa} \cite{DBLP:conf/eacl/SegantiFSSA21} uses an unconventional two-BERT-model pipeline for information extraction.

\textbf{X-METRA-ADA} \cite{DBLP:conf/naacl/MhamdiKDBRM21} employs meta-learning for cross-lingual transfer capability enhancement.

\textbf{SSDM} \cite{DBLP:conf/acl/WuWZXCZ022} proposes a siamese semantic disentanglement model to separate syntax knowledge across languages.

\textbf{LaBSE} \cite{DBLP:conf/acl/FengYCA022} is a BERT-based sentence embedding model supporting 109 languages.

\textbf{DAP} \cite{DBLP:conf/acl/LiHZDLHJWDZ23} integrates sentence-level and token-level dual-alignment for cross-lingual pre-training.

\textbf{SOGO\(_{\text{cos}}\)} \cite{DBLP:conf/emnlp/ZhuCHCZ23} employs saliency-based substitution and a novel token-level alignment strategy for cross-lingual spoken language understanding.

\textbf{X-STA} \cite{DBLP:conf/emnlp/Cao0T0Z23} leverages an attentive teacher, gradient disentangled knowledge sharing, and multi-granularity semantic alignment for cross-lingual machine reading comprehension.

\subsection{General Experimental Results}
\label{sec:general_experimental_results}

\noindent{\bf Sequence Classification:}
The results for sequence classification are presented in Table~\ref{text_classification_general_result} for text classification and Table~\ref{sentiment_analysis_general_result} for sentiment analysis. We observe that:
(1) Our approach outperforms strong baselines and nearly reaches state-of-the-art performance for each task.
(2) The performance of the text classification task is significantly improved by leveraging the Domino Cluster to select appropriate candidates and injecting pseudo semantic knowledge into critical components of the sentences. We achieve an average accuracy of 79.3\%, with a particularly notable improvement for French, where accuracy increases by approximately 10\% (from 74.5\% to 84.4\%) compared to the method proposed by \cite{DBLP:conf/ijcai/QinN0C20}.
(3) In the sentiment analysis task, our UniPSDA model boosts the average macro F1 score to 86.0, as shown in Table~\ref{sentiment_analysis_general_result}. This represents the best reported result for cross-lingual sentiment analysis on the OpeNER dataset.

\begin{table}[tb]
\centering
\small
\resizebox{\linewidth}{!}{
\begin{tabular}{c|ccc|c}
\toprule
\textbf{Model}                                               & \textbf{en}   & \textbf{eu}   & \textbf{ca}   & \textbf{Average} \\ \midrule
BLSE                                                                          & 86.1                           & 88.5                           & 73.9                           & 82.8$_{(\pm 0.4)}$                              \\
m-BERT                                                                        & 89.9                           & 87.9                           & 75.2                           & 84.3$_{(\pm 0.2)}$                              \\
mUSE                                                                          &  88.7    &  90.0    &  75.1    &  84.6$_{(\pm 0.3)}$       \\
XLM-R                                                                         &  87.9    &  87.6    &  71.7    &  82.4$_{(\pm 0.2)}$       \\
DAP                                                                           &  90.5    &  91.7    &  74.9    &  85.7$_{(\pm 0.1)}$       \\
SOGO$_{cos}$                                                                           &   90.1 & 91.2 & 75.0 & 85.4$_{(\pm 0.1)}$       \\
X-STA                                                                           &  90.4 & 90.1 & 74.9 & 85.1$_{(\pm 0.1)}$       \\
CoSDA-ML                                                                      & 90.4                           & 91.6                           & 74.6                           & 85.3$_{(\pm 0.5)}$                              \\ \midrule
\rowcolor[HTML]{C0C0C0}\textbf{UniPSDA} & \textbf{90.7} & \textbf{91.9} & \textbf{75.3} & \textbf{86.0}$_{(\pm 0.1)}$    \\ \bottomrule
\end{tabular}}
\caption{General experimental results of sentiment analysis in terms of macro F1 (\%) and baselines on the OpeNER dataset.}
\label{sentiment_analysis_general_result}
\end{table}

\begin{table}[tb]
\centering
\small
\resizebox{\linewidth}{!}{
\begin{tabular}{c|ccc|c}
\toprule
\textbf{Model}                                               & \textbf{en}   & \textbf{zh}   & \textbf{ar}   & \textbf{Average} \\ \midrule
m-BERT                                                                        & 57.1                           & 44.8                           & 21.5                           & 41.1$_{(\pm 0.5)}$                              \\
XLM-R                                                                         &  54.9    &  45.2    &  21.7    &  40.6$_{(\pm 0.4)}$       \\
LaBSE                                                                         &  55.1    &  45.8    &  26.6    &  42.5$_{(\pm 0.2)}$       \\
HERBERTa                                                                         &  54.5    &  45.7    &  26.4    &  42.2$_{(\pm 0.3)}$       \\
CoSDA-ML                                                                      & 58.3                           & 45.5                           & 20.4                           & 41.4$_{(\pm 0.5)}$                              \\
DAP                                                                           &  59.0    &  46.2    &  26.2    &  43.8$_{(\pm 0.2)}$       \\
SOGO$_{cos}$                                                                           &   58.7 & 46.2 & 26.5 & 43.8$_{(\pm 0.3)}$       \\
X-STA                                                                           &  57.9 & 45.7 & 26.2 & 43.3$_{(\pm 0.1)}$       \\
CCCAR                                                                         & 56.6                           & 43.9                           & 18.3                           & 39.6$_{(\pm 0.3)}$                              \\ \midrule
\rowcolor[HTML]{C0C0C0}\textbf{UniPSDA} & \textbf{59.1} & \textbf{46.3} & \textbf{26.8} & \textbf{44.1}$_{(\pm 0.1)}$    \\ \bottomrule
\end{tabular}}
\caption{General experimental results of information extraction in terms of micro F1 (\%) and baselines on the ACE2005 dataset. }
\label{information_extraction_general_result}
\end{table}



\textbf{Information Extraction:}
The results for information extraction are shown in Table~\ref{information_extraction_general_result}. The findings demonstrate that:
(1) Our methodology is effective for the relation extraction task, achieving more accurate cross-lingual representations as evidenced by higher macro F1 scores compared to prior work.
(2) The enhanced focus on acquiring pertinent cross-lingual knowledge regarding crucial sentence components has led to a solid average F1 score of approximately 44.1 in our experiments, marking an improvement of 2.7.

\textbf{Question Answering:}
Table~\ref{question_answering_general_result} presents the performance of our question answering framework. The results suggest that:
(1) Despite the zero-shot experimental setup, the pseudo data augmentation mechanism employed by our framework demonstrates a robust transfer capability. This translates to effective performance on the BiPaR dataset, with our work producing more accurate representations than most of the baselines.
(2) The scores obtained in the two languages evaluated affirm the efficacy of UniPSDA. However, our F1 scores for English are lower than those achieved by SSDM \cite{DBLP:conf/acl/WuWZXCZ022} and DAP \cite{DBLP:conf/acl/LiHZDLHJWDZ23}. This discrepancy can be attributed to the fact that SSDM and DAP utilize specific parallel data for training, which was not the case in our approach.

\begin{table}[t]
\centering
\small
\resizebox{\linewidth}{!}{
\begin{tabular}{cccccc}
\toprule
\multirow{2}{*}{\textbf{Model}}                                                & \multicolumn{2}{c}{\textbf{en}}                                               & \multicolumn{2}{c}{\textbf{zh}}                                                             & \multirow{2}{*}{\textbf{Average}} \\ \cmidrule{2-5}
                                                                               & \textbf{EM}                                  & \textbf{F1}                    & \textbf{EM}                                  & \textbf{F1}                                  &                                   \\ \midrule
m-BERT                                                                         & 31.9                                         & 44.3                           & 23.5                                         & 26.1                                         & 31.5$_{(\pm 0.6)}$                              \\
XLM-R                                                                          & 32.3                                         & 45.0                           & 23.3                                         & 26.2                                         & 31.7$_{(\pm 0.3)}$                              \\
LaBSE                                                                          & 33.7                                         & 43.4                           & 24.2                                         & 25.7                                         & 31.8$_{(\pm 0.5)}$                              \\
CoSDA-ML                                                                       & 34.2                                         & 44.5                           & 24.7                                         & 25.9                                         & 32.3$_{(\pm 0.2)}$                              \\
X-METRA-ADA                                                                    & 33.9                                         & 44.9                           & 23.8                                         & 26.8                                         & 32.4$_{(\pm 0.1)}$                              \\
SSDM                                                                           & 34.1                                         & 45.8                           & 24.1                                         & 26.2                                         & 32.6$_{(\pm 0.2)}$                              \\
DAP                                                                            & 34.3                                         & \textbf{45.9} & 24.6                                         & 27.1                                         & 33.0$_{(\pm 0.3)}$                              \\ 
SOGO$_{cos}$                                                                            & 34.2 & 45.5 & 24.7 & 27.1 & 32.9$_{(\pm 0.2)}$                              \\ 
X-STA                                                                            & 33.8 & 44.9 & 24.2 & 26.7 & 32.4$_{(\pm 0.2)}$                              \\ 
\midrule
\rowcolor[HTML]{C0C0C0} \textbf{UniPSDA} & \textbf{34.4} & 45.7            & \textbf{25.0} & \textbf{27.3} & \textbf{33.1}$_{(\pm 0.2)}$                              \\ \bottomrule
\end{tabular}}
\caption{General experimental results of question answering and baselines on the BiPaR dataset.}
\vspace{-0.2cm}
\label{question_answering_general_result}
\end{table}





\section{Detailed Analysis}
\subsection{Ablation Study}
\label{sec:ablation_study}

In our ablation study, we independently remove key components—namely, the Domino Unsupervised Cluster module and the Pseudo Semantic Data Augmentation module—to evaluate their individual contributions to the framework’s performance.
The results of the ablation experiments are presented in Table~\ref{ablation_study_result}. We draw two main conclusions:
(1) The Domino Unsupervised Cluster module is crucial for generating precise representations, while the Pseudo Semantic Data Augmentation module significantly enhances the model's performance by providing additional cross-lingual information.
(2) The absence of the cluster module leads to a noticeable decline in performance across all downstream tasks. Specifically, in text classification, accuracy falls by 4.4\% (from 79.3\% to 74.9\%). This indicates that clustering based on semantic embeddings is more beneficial to the model than clustering based on shallow string representations. The removal of the Pseudo Semantic Data Augmentation module also results in a marked decrease in performance due to the lack of cross-lingual knowledge.

\begin{table}[tb]
\centering
\small
\vspace{+0.2cm}
\resizebox{\linewidth}{!}{
\begin{tabular}{l|cccc}
\toprule
\multicolumn{1}{c|}{\textbf{Model}} & \textbf{OpeNER} & \textbf{MLDoc} & \textbf{ACE05}            & \textbf{BiPaR}              \\ \midrule
\multicolumn{1}{c|}{UniPSDA}       & 86.0            & 79.3           &  44.1 & 45.7 \\ \midrule
-Dom. Unsup.                 & 85.8            & 74.9           & 41.9                        & 45.1                        \\ \midrule
-Pse. Seman.               & 85.1            & 73.7           & 41.1                        & 44.8                        \\
-Aff. Regul.                    & 84.2            & 71.1           & 40.9                        & 44.0                        \\ \bottomrule
\end{tabular}}
\caption{Ablation study of our work on four datasets. ``-'' means returning to the original setting.}
\label{ablation_study_result}
\end{table}

\subsection{Influence of Domino Unsupervised Cluster}

\begin{figure}[tb]
\centering
\includegraphics[width=7cm,height=5cm]{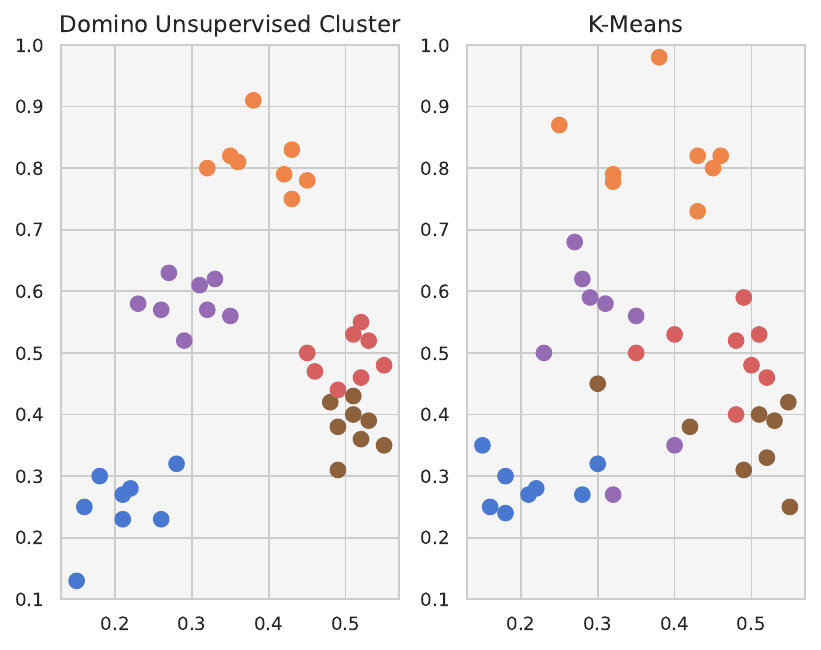} 
\caption{The words representations of five different semantics after t-SNE dimensional reduction.}
\label{influence_of_domino_cluster}
\end{figure}

We employ t-SNE \cite{van2008visualizing} to project the high-dimensional word representations into a two-dimensional space, facilitating the visualization of the embeddings. The resulting plots compare word embeddings clustered by the Domino Unsupervised Cluster and the naive K-Means algorithm. As depicted in Figure~\ref{influence_of_domino_cluster}, the domino unsupervised cluster gathers similar word embeddings more compactly, whereas the naive K-Means approach results in a more diffuse distribution of similar word embeddings.

\subsection{The Influence of Pseudo Semantic Data Augmentation}
\label{sec:influence_pseudo_semantic}

\begin{figure}[tb]
\centering
\includegraphics[width=7cm,height=5.5cm]{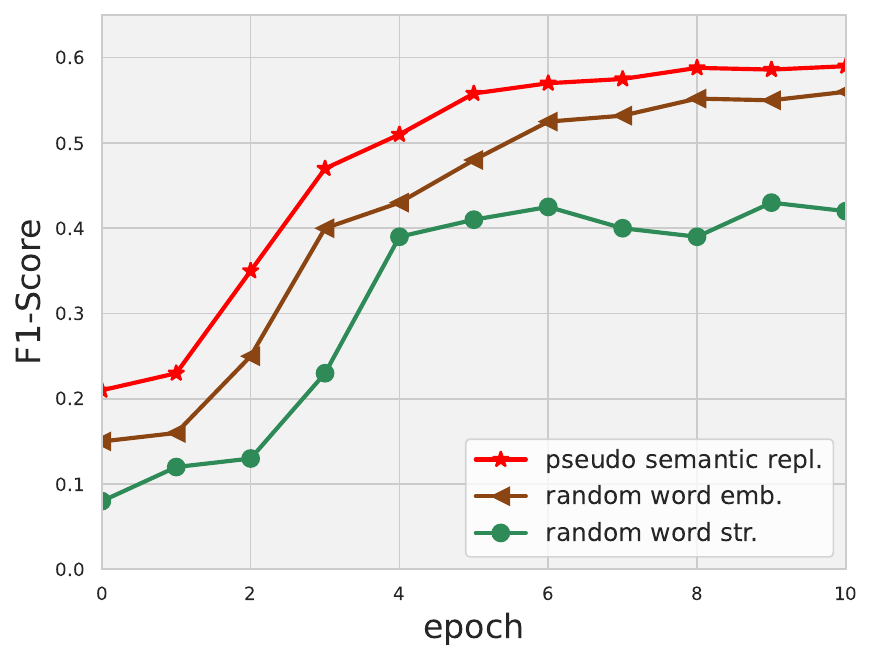} 
\caption{Results comparison of different data augmentation skills in Pseudo Semantic Data Augmentation module.}
\label{influence_of_pseudo_semantic}
\end{figure}

To examine the effect of Pseudo Semantic Data Augmentation on the information extraction task, we experiment with three distinct replacement strategies on the English test set. The comparative results are illustrated in Figure~\ref{influence_of_pseudo_semantic}.

Observations indicate that replacements using random word strings or random word embeddings are less effective than those leveraging pseudo semantic methods. The pseudo semantic data augmentation approach demonstrates a superior ability to preserve the semantic integrity of sentences, leading to more meaningful augmentations and potentially better model performance.


\section{Related Work}
\label{sec:related_work}

\subsection{Cross-Lingual Pre-trained Models}
Recent cross-lingual pre-trained language models (PLMs) can be categorized into two groups:

\begin{enumerate}
  \item \textbf{Monolingual Training Data Models:} Multilingual BERT (m-BERT) \cite{DBLP:conf/naacl/DevlinCLT19} and XLM-R \cite{DBLP:conf/acl/ConneauKGCWGGOZ20} utilize monolingual corpora for training with a masked language modeling task.
  \item \textbf{Multilingual Training Data Models:} Extensions of XLM-R by \citet{DBLP:conf/aaai/JiangLCD22,DBLP:conf/acl/HammerlL022,DBLP:conf/acl/ChiH0MZSBSMHW22,DBLP:conf/lrec/BarbieriAC22} demonstrate improvements with high-quality static embedding alignments. Tools facilitating bilingual language alignment \cite{DBLP:conf/nips/TranTLG20,DBLP:conf/acl/Chi0ZHMHW20,DBLP:conf/aaai/YangMZWL020,DBLP:conf/naacl/SchusterRBG19} enable the learning of additional languages. These models often depend on parallel data and alignment tools to enrich the corpus diversity.
\end{enumerate}

\subsection{Cross-Lingual Data Augmentations}
Cross-lingual data augmentation approaches are typically divided into:

\begin{enumerate}
  \item \textbf{Supervised Data Augmentation:} Methods such as CoSDA-ML \cite{DBLP:conf/ijcai/QinN0C20} and MulDA \cite{DBLP:conf/acl/LiuDBJSM20} utilize parallel corpora to integrate knowledge from other languages. \citet{DBLP:conf/acl/0010ZFXM20} employ parallel language alignments for shared representational spaces.
  
  \item \textbf{Unsupervised Data Augmentation:} Techniques like adversarial training and cross-lingual sample generation are employed by \citet{DBLP:conf/emnlp/RiabiSKSSS21,DBLP:conf/acl/BariMJ20,DBLP:conf/acl/0010ZFXM20,DBLP:conf/emnlp/GuoSPGX0J21} to improve multilingual model performance. \citet{DBLP:conf/acl/NishikawaRT21} use back translation for enhancing word embeddings, while \citet{DBLP:conf/coling/0004HD22} replace words based on a probabilistic distribution. \citet{DBLP:conf/acl/ChenKNGZOM20} focus on sentence selection from low-resource languages. These models tend to prioritize surface string variations, often overlooking the rich, context-aware semantics.
\end{enumerate}

We address this limitation by incorporating global cross-lingual semantics into monolingual training data, thereby enriching the diversity of language knowledge.

\section{Conclusion}
\label{sec:conclusion}

In this work, we introduce UniPSDA, an unsupervised data augmentation mechanism that leverages semantic embeddings to enrich cross-lingual natural language understanding (NLU) tasks with diverse linguistic information. The Domino Unsupervised Cluster module identifies semantically similar cross-lingual content, while the Pseudo Semantic Data Augmentation module injects context-aware semantics into the training corpus. Furthermore, affinity regularization serves to minimize the representational gap between original and augmented sentences. Through extensive experimentation, our methods demonstrate superior performance relative to other strong baselines, underscoring their effectiveness in enhancing cross-lingual NLU.

\section*{Acknowledgements}
We would like to thank anonymous reviewers for their valuable comments. This work was supported in part by National Key R\&D Program of China (No. 2022ZD0120302) and Alibaba Group through Alibaba Research Intern Program.

\section*{Bibliographical References}\label{sec:reference}

\bibliographystyle{lrec-coling2024-natbib}
\bibliography{languageresource}


\end{document}